\documentclass[letterpaper, 10 pt, conference]{ieeeconf}  

\IEEEoverridecommandlockouts                              

\usepackage[utf8]{inputenc}
\usepackage{graphicx}
\usepackage{epstopdf}
\usepackage{amsmath}
\usepackage{amssymb}
\usepackage{multirow}
\usepackage{pbox}
\usepackage{algorithm}
\usepackage{algorithmic}
\usepackage{booktabs}
\usepackage{bm}
\usepackage{url}
\usepackage{esvect}
\usepackage{mathrsfs}

\usepackage{amsthm}
\usepackage{amsfonts}
\usepackage[compact]{titlesec}
\titlespacing{\section}{5pt}{5pt}{5pt}
\titlespacing{\subsection}{5pt}{5pt}{5pt}
\usepackage{cases}

\usepackage[normalem]{ulem}
\newcommand\redsout{\bgroup\markoverwith{\textcolor{red}{\rule[0.5ex]{2pt}{0.4pt}}}\ULon}

\usepackage[caption=false,font=footnotesize,labelfont=sf,textfont=sf]{subfig}
\usepackage{stfloats}

\usepackage[dvipsnames]{xcolor}

\newtheorem{problem}{Problem}

\theoremstyle{definition}

\title{\LARGE \bf A Conformal Mapping-based Framework for Robot-to-Robot and Sim-to-Real Transfer Learning}

\author{Shijie Gao and Nicola Bezzo
\thanks{Shijie Gao, and Nicola Bezzo are with the Charles L. Brown Department of Electrical and Computer Engineering, and Link Lab, University of Virginia, Charlottesville, VA 22904, USA. Email:{\tt\small \{sg9dn, nb6be\}@virginia.edu}}%
}

\renewcommand{\baselinestretch}{0.96}

\begin{document}

\maketitle
\thispagestyle{empty}
\pagestyle{empty}


\begin{abstract}
This paper presents a novel method for transferring motion planning and control policies between a teacher and a learner robot. With this work, we propose to reduce the sim-to-real gap, transfer knowledge designed for a specific system into a different robot, and compensate for system aging and failures. To solve this problem we introduce a Schwarz–Christoffel mapping-based method to geometrically stretch and fit the control inputs from the teacher into the learner command space. We also propose a method based on primitive motion generation to create motion plans and control inputs compatible with the learner's capabilities. Our approach is validated with simulations and experiments with different robotic systems navigating occluding environments.
\end{abstract}

\section{Introduction}
Robotic applications are typically built considering specific systems in mind. For example, popular motion planning methods (e.g., artificial potential field \cite{potential_field}, A* \cite{A_star_alg}, probabilistic techniques \cite{prob_technique}) and control methods (e.g., MPC, PID \cite{control_citation}) require fine tuning and knowledge about system model dynamics in order to be fully leveraged and obtain a desired performance on a selected platform. We also note that most technologies are developed through simulations which offer a practical and inexpensive mean to create and test the limits and performance of designed algorithms.
Researchers usually spend considerable time and resources to create techniques for specific robotic systems and to adapt them on new systems, as well as to compensate for the simulation-reality gap during deployments on actual vehicles.
Finally, even when a new technique is developed and deployed on a specific robot, it can still need to be adjusted or adapted over time due to mechanical aging, disturbances, and even failures that deprecate and modify the system's original model. In this paper we seek a general framework to transfer and adapt system's performance. As mentioned above the goal of the proposed work is to: 
\begin{itemize}
    \item Reduce the sim-to-real gap allowing a developer to quickly transfer motion planning and control methods onto a real platform.
    \item Transfer knowledge designed for a specific robot onto a different robot.
    \item Compensate for system deterioration/failures by learning quickly the limits and the proper input mapping to continue an operation.
\end{itemize}
All of the aforementioned problems can be simplified and cast as a {\em teacher} transferring knowledge to a {\em learner}.

\begin{figure}[ht!]
  \centering
  \includegraphics[width=0.98\columnwidth]{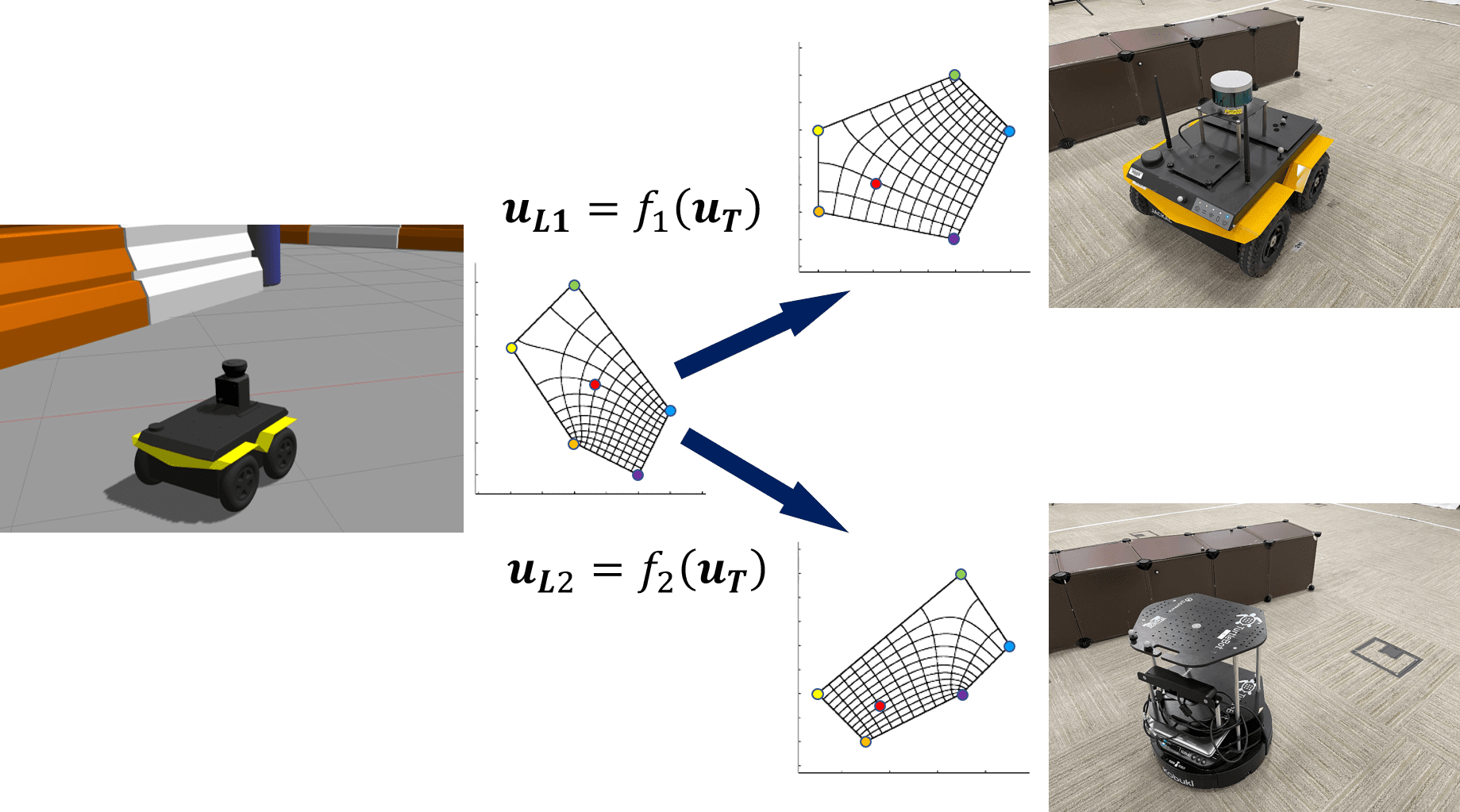}
    \vspace{-8pt}
  \caption{Pictorial representation of the proposed work in which motion planning and control policies are transferred from a teacher simulated vehicle to two vehicles to create the same behavior designed in simulation.}
  \vspace{-18pt}
  \label{fig:INTRO}
\end{figure}

Specifically, to address these problems, in this work we propose a novel method that leverages a variant of Schwarz–Christoffel mapping (SCM) \cite{driscoll2002schwarz} -- a conformal transformation of a simple poly area onto the interior of a rectangle -- to transfer a teacher vehicle's control input sequence to a learner vehicle, as depicted in Fig.~\ref{fig:INTRO}. 
Our proposed method allows the teacher to understand the learner limitations, so that the transferred control input is compatible with the learner capabilities. Finally, once these limitations are extracted, we propose a mechanism to adapt also the teacher motion planning scheme to create paths compatible with the learner constraints. 
To deal with this problem, our scheme leverages an optimized finite horizon primitive motion generation.

 
The main contributions of this work are twofold: 1) a light-weight transfer framework that leverages SCM theory to directly transfers the control input from teacher to learner so that the learner can leverage the teacher's control policy while its own dynamics remain unknown; and 2) a method for adapting the source system's control and path planning policy to the learner. The method constrains the output of the source system's controller and of the path planner so that the transferred motion plan and control input is guaranteed to be compatible with the target system's dynamics.


The rest of the paper is organized as follows: in Section~\ref{sec:relatedWork} we summarize the state-of-the-art approaches for solving sim-to-real problems in the current literature. We formally define the problem in Section~\ref{sec:problemFormulation} while the details of our SCM-based transfer learning framework are presented in Section~\ref{sec:methodology}. The proposed framework is validated with extensive simulations in Section~\ref{sec:simulation} and experiments on real robots in Section~\ref{sec:experiments}. At last, we draw conclusions in Section~\ref{sec:conclusion}.

\section{Related Work} \label{sec:relatedWork}




Transfer learning has been one of the most popular topics in robotics, especially since  machine learning techniques have become widely exploited. The idea behind transfer learning is to migrate the knowledge between similar problems to boost the training process \cite{james2019sim}, take advantage of existing knowledge \cite{devin2017learning}, and reduce the risk of training \cite{fremont2020formal,zhang2020cautious}. Although machine learning approaches have been massively explored, we cannot ignore that they typically require a large amount of data and a lot of effort in training the model. 

The problem of transferring from the simulation to the real world, also known as sim-to-real problem, has gained rising attention recently. The gap between the simulation and the real system exists mainly because either the model is not accurate or the environment factors do not appear in the simulation. The modeling gap can be closed by retraining the pre-trained model in real world \cite{peng2018sim}. Dynamics randomization is another popular solution which aims to cover reality with augmented simulation scenarios \cite{peng2018sim} \cite{tan2018sim}. Other approaches include reducing the costly errors by predicting the blind spots in real environments \cite{ramakrishnan2020blind} and inflating  safety critical regions to reduce the chance of collision \cite{ghosh2019new}. Learning from demonstration is another sub-field of transfer learning in which reinforcement learning is usually getting involved. These types of works typically learn the policy from teacher's examples by approximating the state-action mapping \cite{branavan2009reinforcement}, or by learning the system model \cite{model_learning}. Most of these problems turn into an optimization problem on tuning parameters. Although fewer training demos are desired, it can still take a large amount of data to address the problem. Thus, both the acquisition of data and the tuning process can be challenging when dealing with these types of problems.


To the best of our knowledge, the SCM method proposed in this paper is rarely used in the robotics field. In \cite{notomista2018coverage}, the SCM is leveraged to map the planar motion to the continuous linear motion to solve a coverage control problem for wire-traversing robots. Comparing to the existing works, this paper proposes a light-weight transfer learning framework which does not rely on massive data collection. It is also the first work that exploits the conformal mapping method to directly transferring control inputs between two systems.


\section{Problem Formulation} \label{sec:problemFormulation}




\vspace{-2pt}
The problem behind this work can be cast as a teacher transferring knowledge to a learner vehicle. We assume that the teacher has more capabilities than the learner, meaning that it can achieve all the learner's maneuver but not vice versa. This assumption is suitable for our problem since we are primarily interested in transferring knowledge into a vehicle with degraded capabilities, and as it is easier to create a virtual simulated vehicle with more capabilities than a real vehicle in sim-to-real problems. The learner's dynamics are assumed a black-box model with only access to the inputs and output. The goal is to transition the behavior and control knowledge of the teacher into the learner including adapting the teacher motion planning framework to consider the limitations of the learner. Formally we can define two problems:

\vspace{-3pt}
\begin{problem}\label{problem1}{\bf{\emph{Teacher-Learner Control Transfer:}}}
Given a teacher robot 
with dynamics ${\bm{x_T}(t+1)}{=}f_T(\bm{x_T}(t),\bm{u_T}(t))$ and control law $\bm{u_{T}}{=}g(\bm{x})$, where $\bm{x}$ is the state vector and $\bm{u}$ is the control input, find a policy to map $\bm{u_{T}}$ to a learner input $\bm{u_{L}}$ such that $\bm{x_{L}}(t+1){=}f_L(\bm{x_L}(t),\bm{u_L}(t)){=}\bm{x_{T}}(t+1)$, with $f_L$ unknown.
\end{problem}


\vspace{-7pt}
\begin{problem}\label{problem2}{\bf{\emph{Teacher-Learner  Motion Planning Adaptation:}}} 
Consider a task to navigate from an initial location to a final goal $G$. Assume that the learner's input space $\bm{u_L}\in[{\bm{u_L}}_{min}, {\bm{u_L}}_{max}] \subset [{\bm{u_T}}_{min}, {\bm{u_T}}_{max}]$. Design a motion planning policy $\bm{\pi_{T}}^{L}$ for the teacher that considers the limitations of the learner and such that the computed desired trajectory $\tau$ can be tracked by the learner, i.e., such that $|\bm{x}_L-\bm{x}_{\tau}|\leq \epsilon$ where $\epsilon$ is a maximum allowable deviation threshold.
\end{problem}

\vspace{-3pt}
\section{Methodology}
\label{sec:methodology}
\vspace{-1pt}

\begin{figure}[h]
  \vspace{-10pt}
  \centering
  \includegraphics[width=.85\columnwidth]{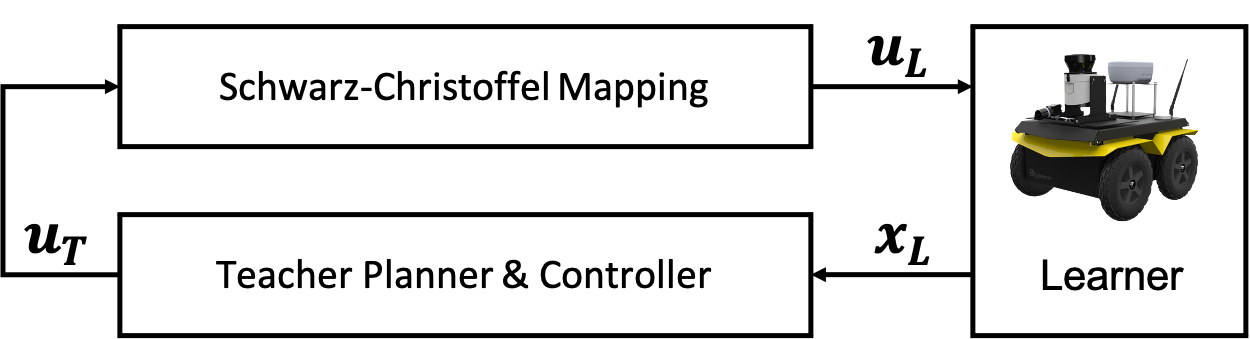}
  \vspace{-4pt}
  \caption{The architecture of the proposed transfer learning process.}
  \label{fig:fullProcessDiagram}
\end{figure}

 
Problem \ref{problem1} is solved by leveraging SCM to comformally map between the teacher's and the learner's command domains. Problem \ref{problem2} is addressed by constraining the teacher's control and planning policy in accordance with the learner's limitation. The block diagram in Fig.~\ref{fig:fullProcessDiagram} shows the architecture of the whole process. 
The  remainder  of  this  section  describes  the details of the components of the proposed approach.

\vspace{-10pt}
\subsection{SCM-based Command Transferring} \label{sec:SCMapping}

As we treat the dynamics of the learner as a black box, it is impossible to build a one-to-one command mapping without running inputs on the learner.
In our work, we propose to use a limited number of teacher commands to characterize the learner's dynamics and then use SCM to find the mapping function between the region on the teacher's command domain and the corresponding region on the learner's side.

\begin{figure}[h!]
\vspace{1pt}
  \centering
  \includegraphics[width=.7\columnwidth]{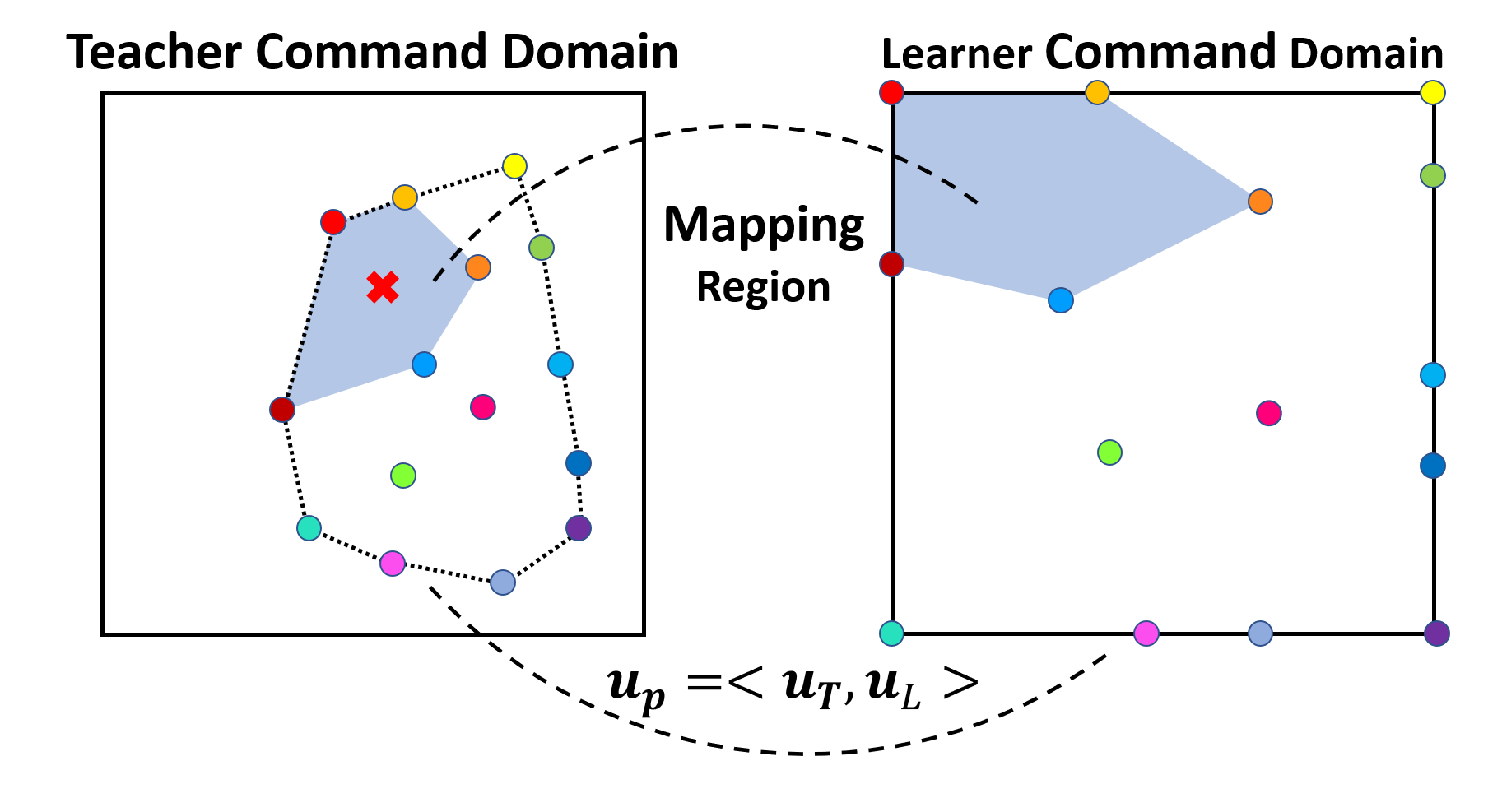}
  \vspace{-18pt}
  \caption{SCM maps the two polygon regions which are constructed by the command pairs around the desired command (red cross on the left).}
  \label{fig:Calibration}
  \vspace{-15pt}
\end{figure}

We use command pairs to characterize the learner's dynamics. The command pair $\bm{u_{p}}{=}\langle\bm{u_T}, \bm{u_L}\rangle$ is a pair of commands which makes the two vehicles produce the same motion (i.e., reach the same pose, speed). Since the dynamics of the teacher are known, by observing the states of the learner before and after executing $\bm{u_L}$, the equivalent teacher's command $\bm{u_T}$ can be retrieved. A group of these command pairs can capture the dynamics of the learner on the teacher command domain. At each control step, the learner uses the teacher's control policy to generate a control input which is the teacher's desired command as if the learner was the teacher. Given a desired teacher's command and several command pairs around it, the region whose vertices are from the command pairs and contains the desired command can be chosen on the teacher side. The corresponding region on the learner command domain is decided automatically by the learner's commands that come from the same command pairs as the teacher's vertices. An example is shown in Fig.~\ref{fig:Calibration}.

\begin{figure}[h]
    \centering
    \includegraphics[width=.9\columnwidth]{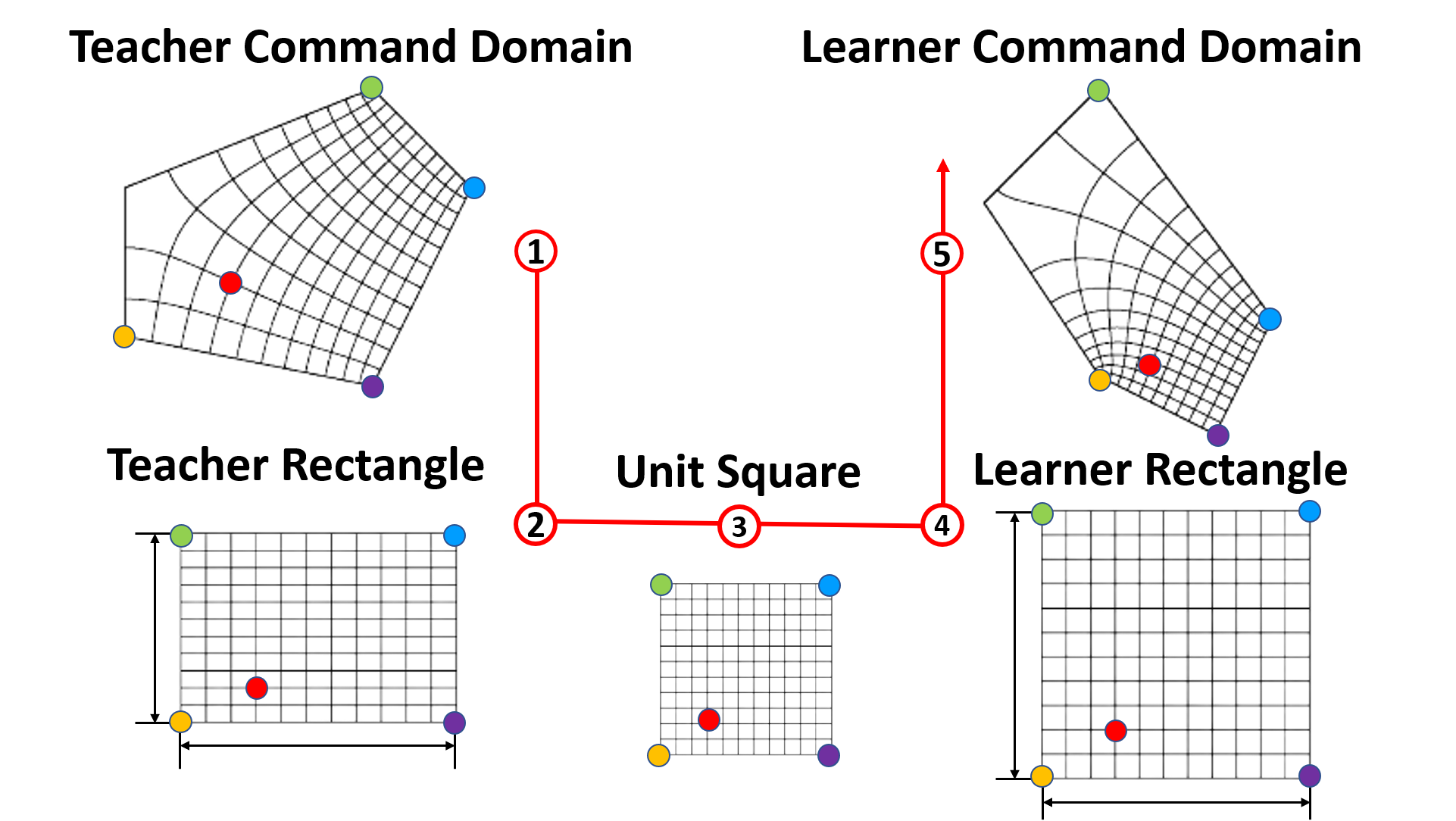}
    \vspace{-10pt}
    \caption{The mapping flow of transferring the desired teacher command to the learner. A unit square is used as an intermediate plane to bridge between rectangle mapping of the two polygons.}
    \label{fig:mappingFlow}
\end{figure}

Once the regions of interest are determined on both teacher's and learner's command space, the transfer problem becomes a problem of finding the mapping function that transfers from an irregular polygon on the teacher's domain to the other polygon on the learner's domain. To solve this problem, first we use SCM to map the two polygons on each side of the command domain onto two rectangles with unique aspect ratios, which are decided by the shape of the mapping area. The reason why we map the two regions onto two different rectangles will appear as we walk through the mapping procedure.   Then, we use a unit square to bridge the two rectangles so a teacher command can be mapped to the learner's domain. Fig.~\ref{fig:mappingFlow} shows the mapping flow.

\vspace{-5pt}
\begin{figure}[h]
    \centering
    \includegraphics[width=.9\columnwidth]{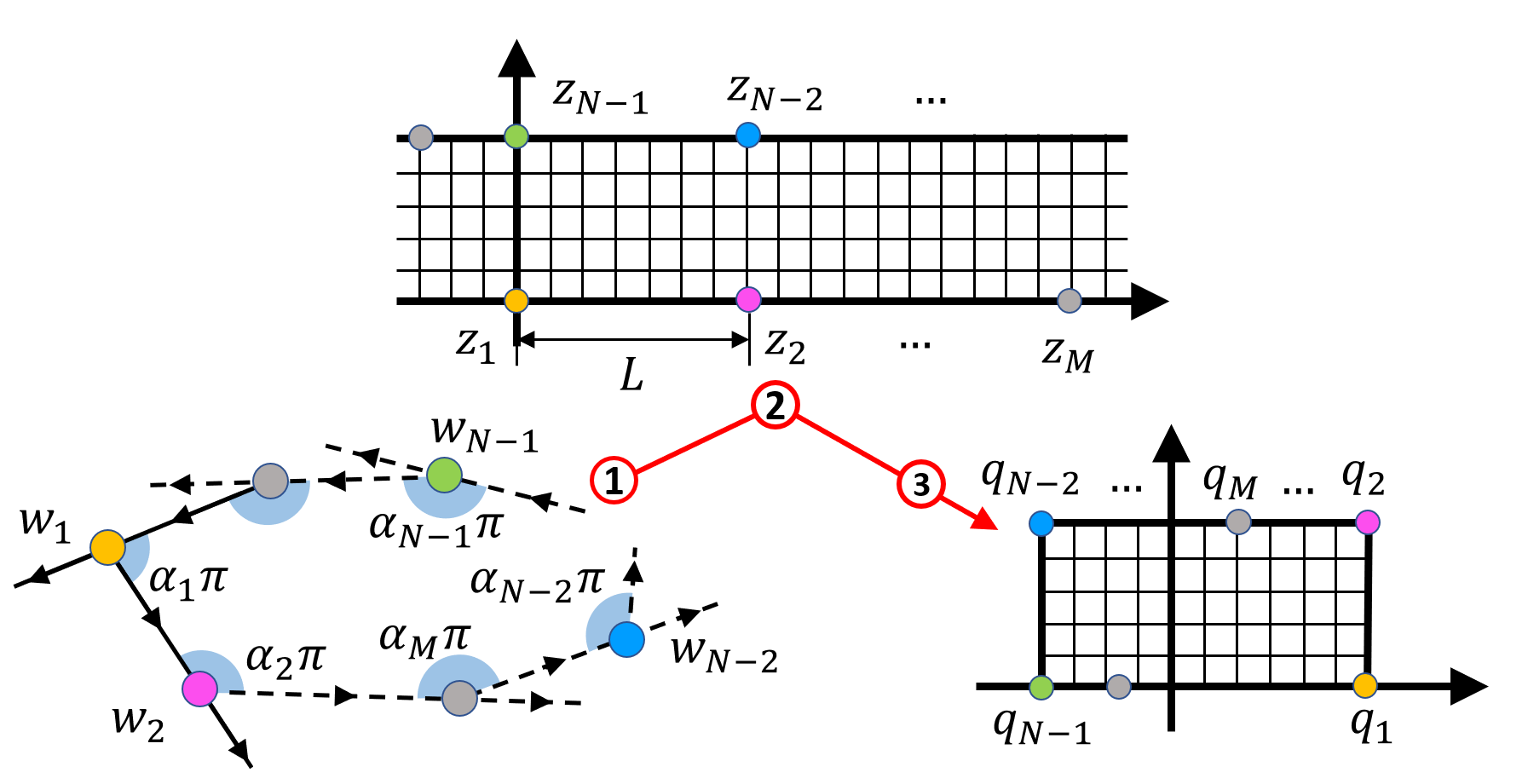}
    \vspace{-10pt}
    \caption{The flow of conformal mapping that maps the polygon to the rectangle while using the bi-infinite strip as the intermediate plane.}
    \label{fig:StripMapping}
\end{figure}

\vspace{-7pt}
Based on the user's preference, multiple command pairs can be selected to build the mapping areas $\Gamma$. For any of these irregular polygons, we can specify four of the vertices in the counterclockwise order to map to the rectangle's corners. These four vertices make $\Gamma$ a generalized quadrilateral. Fig.~\ref{fig:StripMapping} shows an example of this process, where we put the polygon from the teacher command domain onto the extended complex plane.
\vspace{-10pt}

\vspace{-5pt}
As shown in Fig.~\ref{fig:StripMapping}, the vertices of the polygon $w_1, ..., w_N$, $(N{\geq} 4)$ are ordered in counterclockwise and the interior angles $\alpha_1\pi, ..., \alpha_n\pi$ at each of the vertex $w_N$ is defined as the angle that sweeps from the outgoing edge to the incoming edge. The conformal mapping from the polygon $\Gamma$ to the rectangle $\mathbb{Q}$ needs to borrow a bi-infinite strip $\mathbb{S}$ as an intermediate plane. The SCM function that maps the points on the boundary of the strip $\mathbb{S}$
to the vertices of the polygon is given by:

\vspace{-10pt}
\begin{equation} \label{eql:strip2polygon}
    w =  f_{\mathbb{S}}^{\Gamma}(z) = A\int_{0}^{z}\prod_{j=0}^{N} f_{j}(z)dz +C
    \vspace{-5pt}
\end{equation}
where $A$ and $C$ are complex constants that rotate, translate and scale the polygon and are determined by its shape and location. Each factor $f_j$ sends a point on the boundary of the strip to a corner of the polygon while preserving its interior angles. The factor $f_j$ is a piecewise function which is defined by: 
\vspace{-10pt}
\begin{numcases}{f_j(z) {=}} \label{eql:FunctionFactors}
    e^{\frac{1}{2}(\theta_+ - \theta_-)z} & $j {=} 0$, \nonumber \\
    \{-i \cdot \sinh[{\textstyle\frac{\pi}{2}}(z - z_j)]\}^{\alpha_j} & $1\leq j \leq M$,\\
    \{-i \cdot \sinh[-{\textstyle\frac{\pi}{2}}(z - z_j)]\}^{\alpha_j} & $M+1\leq j \leq N$, \nonumber
\end{numcases}
where $M$ is the number points on the bottom side of the strip. $\theta_+$ and $\theta_-$ denote the desired divergence angles at $+\infty$ and $-\infty$, which are $\theta_+ {=} \theta_- {=} \pi$ in our case. 

By leveraging the Jacobi elliptic of the first kind \cite{Byrd1971Handbook}, the SCM mapping $f_{\mathbb{Q}}^{\mathbb{S}}$ from the rectangle $\mathbb{Q}$ to the bi-infinite strip $\mathbb{S}$ can be defined by:
\vspace{-4pt}
\begin{equation} \label{eql:Jacobielliptic}
    z = f_{\mathbb{Q}}^{\mathbb{S}}(q) = \frac{1}{\pi}\cdot \ln(\sin(q|m))
    \vspace{-3pt}
\end{equation}
where $q$ is the point on regular rectangle and $m$ is the modulus of the Jacobi elliptic that is decided by $q$. The details of this conformal mapping can be found in \cite{driscoll2002schwarz}. With Eqs. \eqref{eql:strip2polygon} and \eqref{eql:Jacobielliptic}, a mapping function from the generalized quadrilateral can be obtained. In order to explicitly solve \eqref{eql:strip2polygon}, there are three parameters $z_k$ that must be specified. For ease of computation, for example, we can fix $z_1 = 0$, $z_2 = L$, $z_{N-1} = i$, and $z_{N-2} = L+i$. The parameter $L$ here is linked to the conformal modulus $m$.

While the angles of the polygon are computed with \eqref{eql:strip2polygon} and \eqref{eql:FunctionFactors}, we need to find where the pre-vertices lie on the boundary of the strip to keep the length for each edge of polygon. This problem is known as the parameter problem in SCM \cite{driscoll2002schwarz}. Since we already fix $z_{1} = 0$, in \eqref{eql:strip2polygon} the translation parameter is set to be $C = 0$. Hence, solving \eqref{eql:strip2polygon} is equal to solving:
\vspace{-8pt}
\begin{equation} \label{eql:strip2polygon_simplified}
        w_k = A\int_{}^{z_k}\prod_{j=0}^{N} f_{j}(z)dz, \quad k = 1,2,3,\dots,N
        \vspace{-4pt}
\end{equation}

In \eqref{eql:strip2polygon_simplified}, the scalar $A$ can be eliminated by the ratio of the adjacent sides length of the polygon:
\begin{equation} \label{eql:ratio}
    \frac{w_{k+1}-w_k}{w_2-w_1} {=} \frac{\int_{z_k}^{z_{k+1}}\prod_{j=0}^{N} f_{j}(z)dz}{\int_{z_1}^{z_2}\prod_{j=0}^{N} f_{j}(z)dz}, \:  k {=} 2,3,\dots,N-2
    \vspace{-5pt}
\end{equation}
Let 
\vspace{-5pt}
\begin{equation}
    I_k = \Big| \int_{z_k}^{z_{k+1}}\prod_{j=0}^{N} f_{j}(z)dz \Big|, \quad k = 1,2,\dots,N-2
    \vspace{-3pt}
\end{equation}
Then \eqref{eql:ratio} can be rewritten as:
\vspace{-2pt}
\begin{equation} \label{eql:final}
    I_k = I_1 \cdot \frac{w_{k+1}-w_k}{w_2-w_1}, \quad k = 2,3,\dots, N-1
    \vspace{-3pt}
\end{equation}

To this end, \eqref{eql:final} leaves us $N-3$ conditions and the unknown parameters of \eqref{eql:strip2polygon_simplified} are $z_k\, (k = 1,2,\dots,N-3)$ which is exactly the number of the side length conditions given by \eqref{eql:final} 
. We can get the complex constant $A$ by:

\begin{equation} 
    A = \frac{w_2-w_1}{\int_{z_1}^{z_2}\prod_{j=0}^{N} f_{j}(z)dz}. 
\vspace{-1pt}
\end{equation}

As we get the conformal mapping function $f_{\mathbb{S}}^{\Gamma}$ from the strip to the generalized quadrilateral, we can compute $ L = z_2- z_1 = {f_{\mathbb{S}}^{\Gamma}}^{-1}(w_2) - 0$. Considering \eqref{eql:Jacobielliptic} which maps the rectangle to the strip, the SCM function that maps the interior and the boundary of the generalized quadrilateral to the rectangle with an unique aspect ratio can be obtained by:
\begin{equation}
    \vspace{-2pt}
    q = f_{SCM}(w) = {f_{\mathbb{Q}}^{\mathbb{S}}}^{-1}({f_{\mathbb{S}}^{\Gamma}}^{-1}(w)).
\end{equation}

As the shape of the rectangle $\mathbb{Q}$ depends on the parameter $L$, the aspect ratio of the rectangle is determined after $L$ is computed. 
This explains why we map the two polygons from teacher and the learner command domains to two different rectangles. Since the dynamics of the teacher and learner are different, the shape of the polygons from the teacher and the learner cannot be identical, and neither are the mapped rectangles. A unit square is borrowed to bridge between the two mapped rectangles resulting in a complete mapping process from teacher to the learner, such that any teacher command that falls in the teacher's mapping area is connected to an image on the learner side.

There are a few points that are worth mentioning: 1) Although we use rectangle SCM and the number of the vertices for a polygon is at least 4 ($N \geq 4$), this mapping-based transferring framework still works for the triangle areas ($N =3$) by leveraging a disk SCM function or an upper half-plane SCM function. 2) 
If the distance between the desired command and the existed closest command pair is smaller than a threshold $ \psi $, it means that the desired motion is very similar to the motion produced by the closest pair. In this case, it is reasonable to skip the mapping procedure and directly use the learner's command from the closest pair. 3) If the command pairs that are used for constructing the mapping polygon are too far from the desired command, some local geometric features between the two domains may not be well captured during mapping. Thus, the number as well as the distribution of the command pairs can affect the mapping performance. More command pairs that cover the learner's command domain well are preferred.

\subsection{Primitive Path Planning}


As the vehicle learns the mapping function, it is also important to know the limitations of the learner so that the teacher's policy can generate the command to plan the motions that are compatible with the learner. This means that we want to find where the command boundary of the learner lies within the teacher command domain. This can be achieved by getting the command pair $\bm{u_p} {=} \langle\bm{u_{T}}(t), \bm{u_{L}}(t)\rangle $ when $\bm{u_{L}}(t) {=}{\bm{u_{L}}}_{max}$. As shown in Fig.~\ref{fig:Calibration}, the teacher's control inputs from these command pairs can build a multi-dimensional convex hull that separates the interior of the convex hull from the rest of the command area. From the teacher's perspective, the boundary of the convex hull indicates the limitations of the learner.
Any of teacher's commands from the interior of the convex hull can be matched with the learner's command, enabling the two vehicles to produce the similar motion with their own commands. However, as it is pointed out at the end of Section \ref{sec:SCMapping}, to obtain better mapping performance, it is recommended to consider additional command pairs inside of the polygon.

We use a trajectory tracking case study to validate our approach. The teacher uses a search-based path planning method to compose a sequence of motion primitives that allows it to drive along the desired path $P$ within a certain bounds. The teacher's input sequence associated to these primitives will be the desired commands for mapping.

\begin{figure}[ht!]
  \centering
  \includegraphics[width=.85\columnwidth]{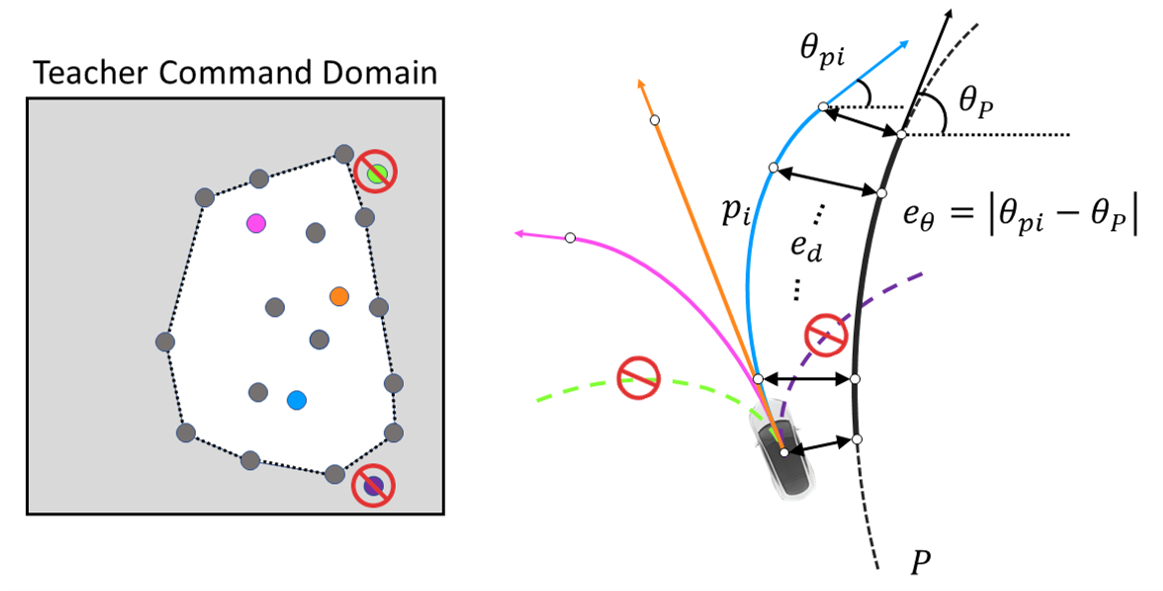}
  \vspace{-8pt}
  \caption{The teacher commands and the corresponding motion primitives are shown on the left while a path planning scenario is shown on the right.}
  \vspace{-7pt}
  \label{fig:PrimitivePathPlanning}
\end{figure}

A motion primitive results from feeding a known sequence of control inputs to the vehicle. 
To build one primitive $p {=} [ {\bm{x_T}}_1,{\bm{x_T}}_2,\dots,{\bm{x_T}}_t ]$, we feed the teacher a sequence of the same control input for a certain amount of time and record its state sequence. 
Following the same procedure, a library of primitives can be built with different teacher's command. In Fig.~\ref{fig:PrimitivePathPlanning}, we show 5 different motion primitives that resulted from 5 different teacher's commands. The one-to-one primitives and the corresponding commands are color coded. The command pairs are shown as the gray points and the white region indicates the capability of the learner.
\vspace{-7pt}

We want to point out that: 1) 
To better adapt to the capability of the learner, only the command which falls inside of the convex hull should be considered. 2) 
The learner can leverage the teacher's motion planner as soon as the convex hull is built. 3)  The convex hull does not need to capture the entire command domain of the learner, it just provides a boundary that make sure the learner is operating within the known capability.



As the path planner searches primitives from the library to use, it evaluates the difference between each of the primitive and the corresponding segment on the desired path. As shown in \eqref{eql:differencePrimitivePathPlanning} and in Fig.~\ref{fig:PrimitivePathPlanning}, the difference is measured by considering both the dynamic time warping (DTW) distance $e_{d}$ and the heading difference $e_{\theta}$ at the end of the primitive:
\begin{equation}
\begin{split}
    \delta_i&= k_d \cdot e_{d} + k_{\theta} \cdot e_{\theta} \\
            &= k_d \cdot DTW(P, p_i) + k_\theta \cdot | (\theta_{P} - \theta_{p_i})|,\\    
    p_{i}^* &= \min_{p_1, ..., p_i} \delta_i. \\
\end{split}
\label{eql:differencePrimitivePathPlanning}
\vspace{-2pt}
\end{equation}
    The two types of differences are weighted by two user-defined gains ($k_d {\geq} 0$, $k_\theta {\geq} 0$). 
A large $k_d$ will force the vehicle to remain close to the trajectory while a large $k_t$ will give the primitives that are parallel to the trajectory a better chance to be chosen. Using this metrics, the planner searches through all the primitives in the library and selects the one with the least difference as the optimal local path plan $p_i^{*}$. The teacher's control input $\bm{u_{T}}^{*}$, which is associated to $p_i^{*}$, is the command that will be mapped to the learner.

After a command sequence is executed, the learner will evaluate the situation and use the planner to generate a new local path and corresponding command sequence. The learner will continue to repeat this planning procedure until it arrives to the destination.

Since the learner has differing dynamics from the teacher, as the learner executes the command sequence to follow the composed path, it may deviate from it. When the learner is in an open area, such deviation is not critical because the command sequence only lasts a short period of time and it can always be corrected by the planner at the next planning step. However, such deviation can compromise the safety of the learner when it maneuvers in a cluttered environment. To provide safety guarantees to the system, we introduce an event triggered mechanism to monitor the learner at runtime. The runtime monitor measures the distance between the learner and the planned path $d_{\hat{e}}$. The re-planning procedure is triggered when $d_{\hat{e}} {>} \epsilon$. The smaller that the threshold $\epsilon$ is, the more conservative the learner behaves. As we discussed, the learner does not need to constantly re-plan if the deviation happens in an open area. Thus, the threshold $\epsilon$ should be dynamically changed to reflect how crowded the surroundings are. In our work, the threshold is defined as:
\begin{equation}
    \epsilon =
    \begin{cases}
    \eta * \min(|| p - o_i||) & i = 1,2, \ldots , N_{o}, \\
    \infty & i = \varnothing, \\
    \end{cases}
\end{equation}
where $N_o$ is the number of obstacles in the learner's field of view, $o_i$ is the position of obstacle $i$, and $\eta$ is a constant.

\section{Simulations} \label{sec:simulation}

\begin{figure*}[b]
  \centering
  \includegraphics[width=1\textwidth]{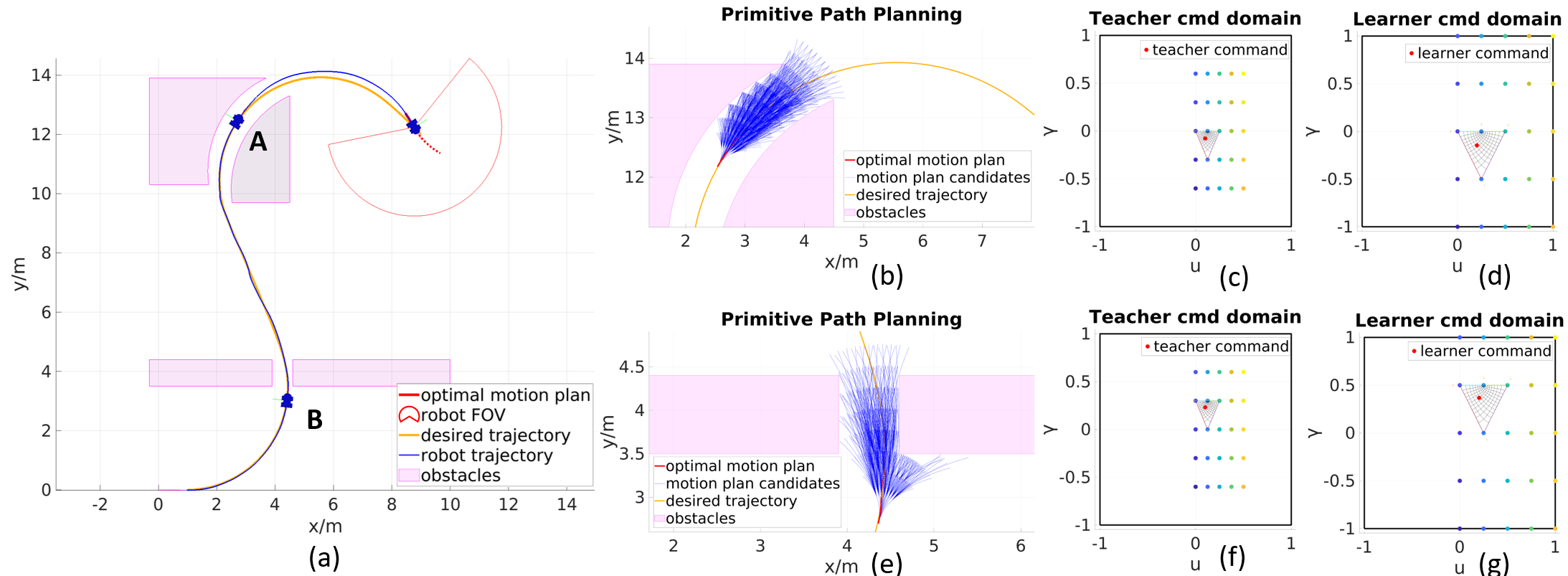}
  \caption{The path following result of the entire simulation is depicted in (a). The local path planning of the SCM mapping results for the robot at position `\textbf{A}' are shown in (b), (c), (d), and the results at position `\textbf{B}' are shown in (e), (f), and (g).}
  \label{fig:simulation_result}
\end{figure*}

\vspace{-5pt}
For the simulations, we created a general case study which, we believe, is rich enough to represent the problems we are dealing with. With the following case studies we demonstrate how, thanks to our approach, a robot can quickly adapt to downgraded dynamics due for example to a failure or system's aging.
In this case, the teacher is a vehicle with full capabilities while the learner is the same vehicle whose dynamics are compromised. For ease of implementation, we consider that both the teacher and the learner have small inertia thus the acceleration period can be neglected (e.g., an electric vehicle). The kinematics for both the teacher and the learner are given by the following bicycle model: 
\begin{align}\label{eql:dynamicmodel}
\begin{split}
    \begin{bmatrix} 
      \dot{x} \\
      \dot{y} \\
      \dot{\theta} 
      \end{bmatrix} = \begin{bmatrix} 
      (v\cdot v_{max}) \cdot \cos\theta \\
      (v\cdot v_{max}) \cdot \sin\theta \\
      \gamma \cdot \gamma_{max}
      \end{bmatrix}, \;\;\;
      \bm{u} &= \begin{bmatrix}
      v\\
      \gamma\\
      \end{bmatrix},
\end{split}
\end{align}
where $v_{max}$ and $\gamma_{max}$ denote the maximum capability on velocity and steering angle of the vehicle. The learner's model is treated as a black box which takes in a control input and produces the updated state of the learner. A Gaussian noise of $G \sim \mathcal{N}(0, 0.1)$ is added to the learner's position to simulate measurement errors. Since the teacher and the learner are the same vehicle, the range of the control inputs for both of the vehicles are same which are $\bm{u} {=} \{v, \gamma \; | \; v\in[0,1], \gamma \in [-1,1]\}$. However, the learner is downgraded so that it can not achieve the same level of performance as the teacher when it is given the same command. In this case study, the maximum velocity $v_{max}$ of the learner is downgraded from $3$~m/s to $1$~m/s while the maximum steering angle $\gamma_{max}$ is downgraded from $\pi/3$~rad/s to $\pi/8$~rad/s. For example, the same control input $v{=}1$ drives the teacher at $3$~m/s while the learner can only drive at $1$~m/s. The learner is asked to follow a "S"-shaped trajectory while navigating through a cluttered environment. 

\begin{figure}[t]
  \centering
  \includegraphics[width=.85\columnwidth]{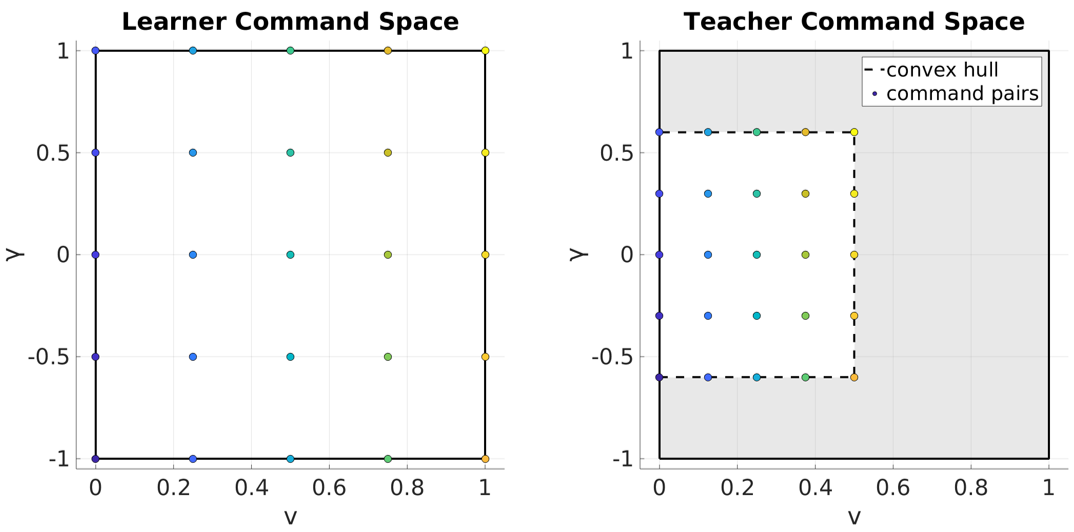}
  \vspace{-10pt}
  \caption{The command pairs are one-to-one color coded across the two command domains.}
  \label{fig:commandPairs}
  \vspace{-15pt}
\end{figure}

Fig.~\ref{fig:simulation_result} shows two snapshots within the time frame of the entire simulation. As the result shows, the learner is able to closely follow the desired trajectory. The learner behaves more conservatively when the obstacles are within the field of view (FOV). In order to obtain the results in Fig.~\ref{fig:simulation_result}, a sequence of $5\times5$ grid commands were fed to the learner. Based on the change of the states before and after executing the command, an equivalent teacher command is retrieved and paired with learner's input. All the command pairs are shown in Fig.~\ref{fig:commandPairs}. The boundary of the commands on teacher's command space marks the limitation of the learner. The learner can map the teacher's command which falls in the boundary to get the learner's control input, and the mapped control input will produce a similar maneuver as the teacher.

\begin{figure}[h]
 \vspace{-5pt}
  \centering
  \includegraphics[width=.85\columnwidth]{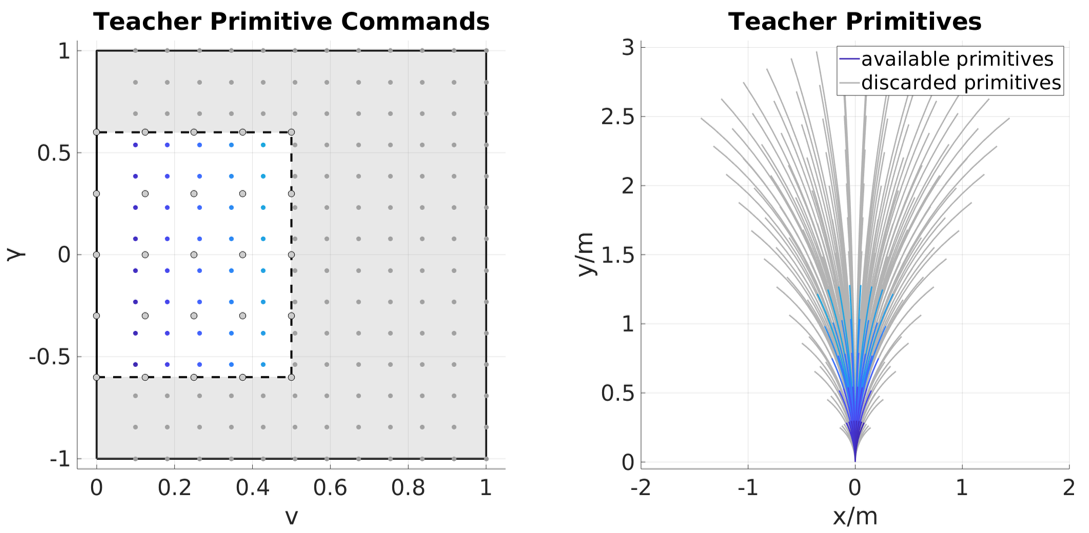}
  \vspace{-10pt}
  \caption{The primitives associated with the small gray commands in shaded area are beyond the limitation of the learner and thus are discarded. The available motion primitives and the associated commands are color coded.}
  \label{fig:primitives}
  \vspace{-8pt}
\end{figure}

Fig.~\ref{fig:primitives} shows all the teacher's motion primitives and the corresponding commands. Each of the primitives are constructed by driving the teacher with a certain control input for $1$ second. The command pairs on the boundary of the convex hull are used to identify if the command for building the motion primitive is within the learner's capability. Among all the 121 motion primitives, 35 of them are preserved after the motion degradation and used for path planning. 

\begin{figure}[h]
\vspace{-10pt}
  \centering
  \includegraphics[width=0.7\columnwidth]{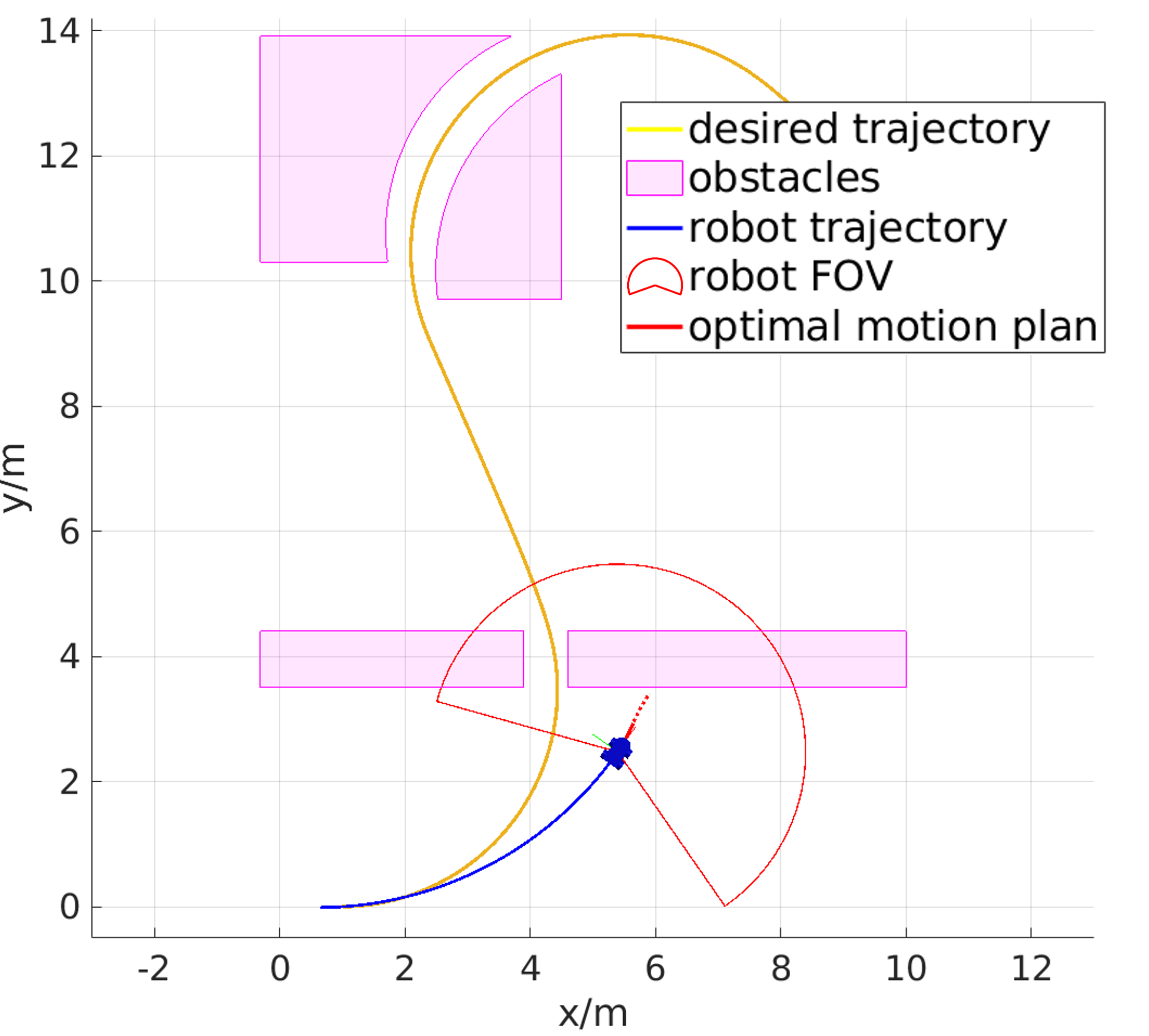}
  \vspace{-11pt}
  \caption{Simulation result for the case in which the downgraded learner is directly given the teacher's commands.}
  \vspace{-10pt}
  \label{fig:simularion_noSCM}
\end{figure}

For the path planner, we set the planning horizon to $s {=} 2$ and the threshold to trigger re-planning as $\eta {=} 0.5$. In Fig.~\ref{fig:simularion_noSCM}, we show the result of the learner driving directly with the teacher's commands without using our proposed approach. As expected, the learner failed because it used commands not adapted to its new dynamics. 

\section{Experiments} \label{sec:experiments}
Our proposed transfer learning approach was validated by a set of experiments in which we transferred the planning and control knowledge of a simulated teacher into two real learner vehicles. The video of all experiments are available in the provided supplemental material. In each of the experiments, we used the same simulated teacher vehicle. The vehicle dynamic model can be approximated to the one showed in the simulation experiments. The maximum velocity $v_{max}$ and the maximum steering angle $\gamma_{max}$ of the teacher were set to be $1.6$~m/s and $\pm 1.2$~rad/s respectively. The proposed method was implemented in MATLAB and we used the MATLAB ROS Toolbox together with Robot Operating System (ROS) to control the vehicles. We used MATLAB Schwarz-Christoffel toolbox \cite{driscoll2005algorithm} for computing the mapping function. The experiments were conducted in the indoor environment and the state of the vehicles are captured by a VICON motion capture system.

\begin{figure}[h]
  \vspace{-10pt}
  \centering
  \includegraphics[width=.9\columnwidth]{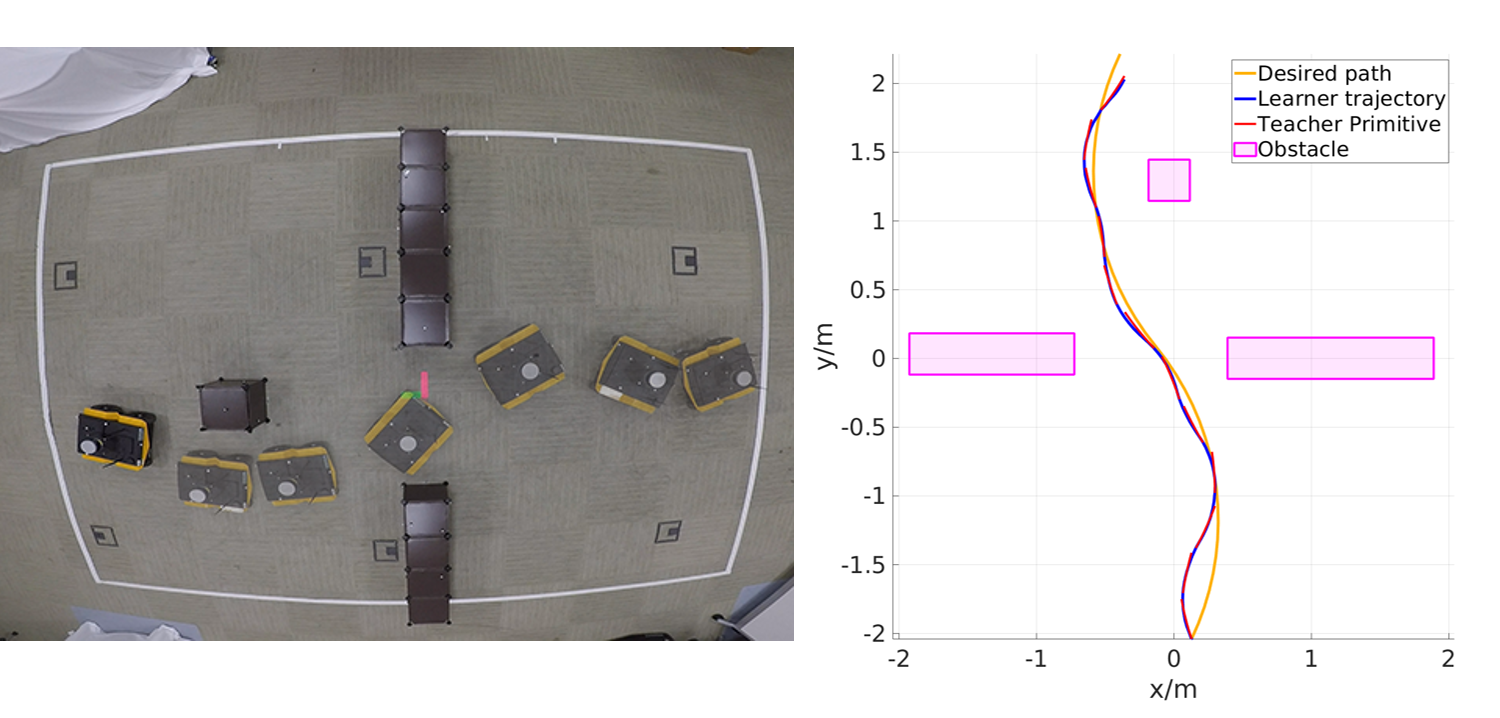}
  \vspace{-10pt}
  \caption{Jackal experiment with SCM.}
  \label{fig:ExpJackalSCM}
\end{figure}

\begin{figure}[h]
\vspace{-17pt}
  \centering
  \includegraphics[width=1\columnwidth]{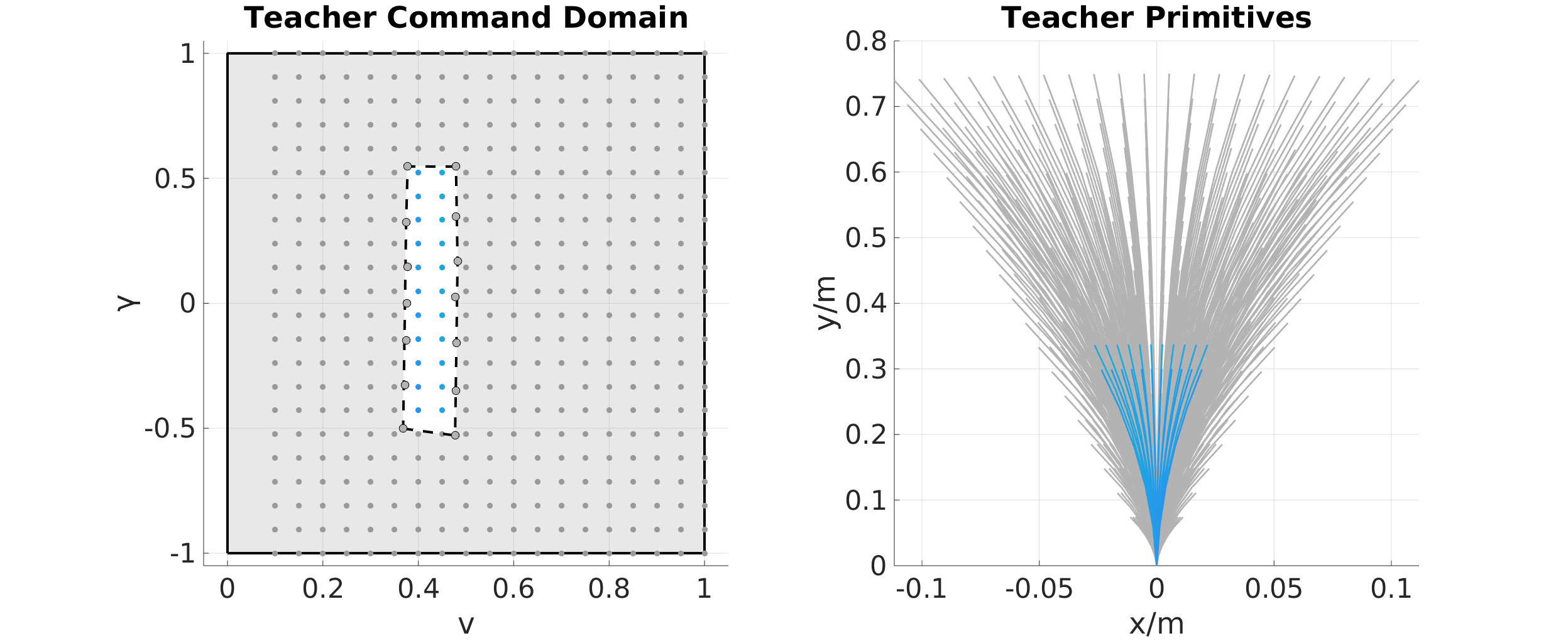}
  \vspace{-20pt}
  \caption{The Jackal's capability is indicated within the white area. The gray points on the dashed boundary are the commands that were tested on the Jackal for extracting the limitations. The blue colored commands on the left create the primitives on the right and are used for mapping to the real UGV.}
  \label{fig:JackalCalibration}
\end{figure}

\begin{figure}[h]
\vspace{-18pt}
  \centering
  \includegraphics[width=.9\columnwidth]{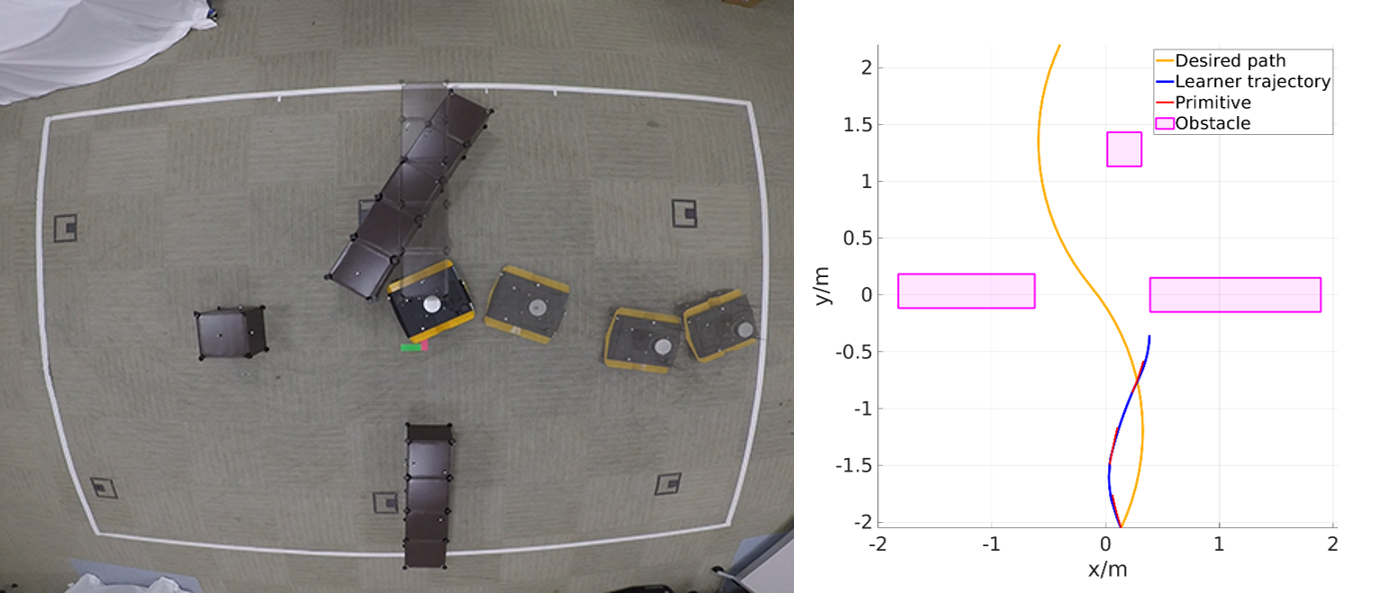}
  \caption{Jackal experiment by directly feeding teacher's command.}
  \vspace{-3pt}
  \label{fig:ExpJackalNoSCM}
\end{figure}

For the first experiment, we asked the learner vehicle to follow an S-shaped path with the initial heading of $\frac{\pi}{4}$ from the desired orientation. As shown in Fig.~\ref{fig:ExpJackalSCM}, a narrow gate and an obstacle was set along the path. Using a Clearpath Jackal UGV as the learner vehicle, we tested its capability by sending certain commands over a period of 1 second, and based on the change to the state, we retrieved the equivalent teacher commands. The command pairs and the teacher's primitives that were used to plan the learner's path are demonstrated in Fig.~\ref{fig:JackalCalibration}. During the tracking mission, the maximum distance between the desired path and the actual trajectory was recorded as $0.1905$~m and the maximum deviation between the actual trajectory and the local motion plan was $0.0293$~m. Considering the vehicle's initial heading is not aligned with the desired path and the size of the vehicle is approximately $0.5$~m$\times0.43$~m$\times0.25$~m, the maximum deviation was negligible. For comparison, the same experiment without the SCM component was performed. 
As expected and as shown in Fig.~\ref{fig:ExpJackalNoSCM}, the learner vehicle collided with the gate and could not continue its task. Additionally, it can be clearly seen that there was a mismatch between the learner's trajectory and the primitive which was given by the path planner. This is also due to the fact that the teacher's control input was not mapped to the learner.

\begin{figure}[h]
  \vspace{-5pt}
  \centering
  \includegraphics[width=1.1\columnwidth]{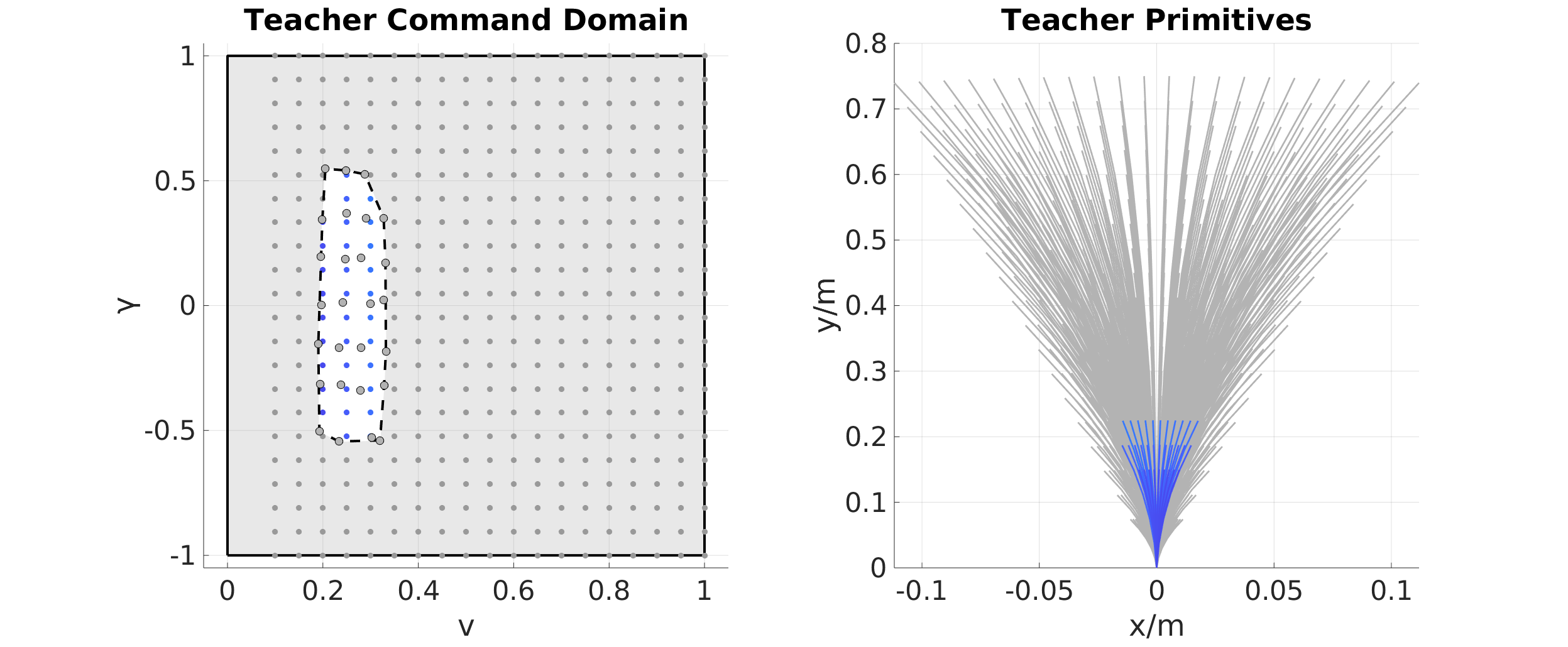}
  \vspace{-15pt}
  \caption{Similar to the Jackal experiment, the turtlebot experiment command pairs and primitives are shown in the figure.}
  \label{fig:TurtleCalibration}
\end{figure}

To show the generalizability of our proposed framework,
similar to the experiment with the Jackal UGV, we performed another experiment with the same settings but this time using a Turtlebot2 as learner. The command pairs and the primitives which were used for learner path planning are shown in Fig.~\ref{fig:TurtleCalibration}. The result shows that with our proposed approach, the Turtlebot2 could adapt the teacher controller and path planner to track the desired path with the maximum deviation of $0.1381$~m. The tracking error between the vehicle's trajectory and the local planned primitive was small within $0.0978$~m as can be noted in the figure in which the blue and the red segments are nearly overlapping throughout the whole process.

\begin{figure}[h]
  \vspace{-20pt}
  \centering
  \includegraphics[width=.9\columnwidth]{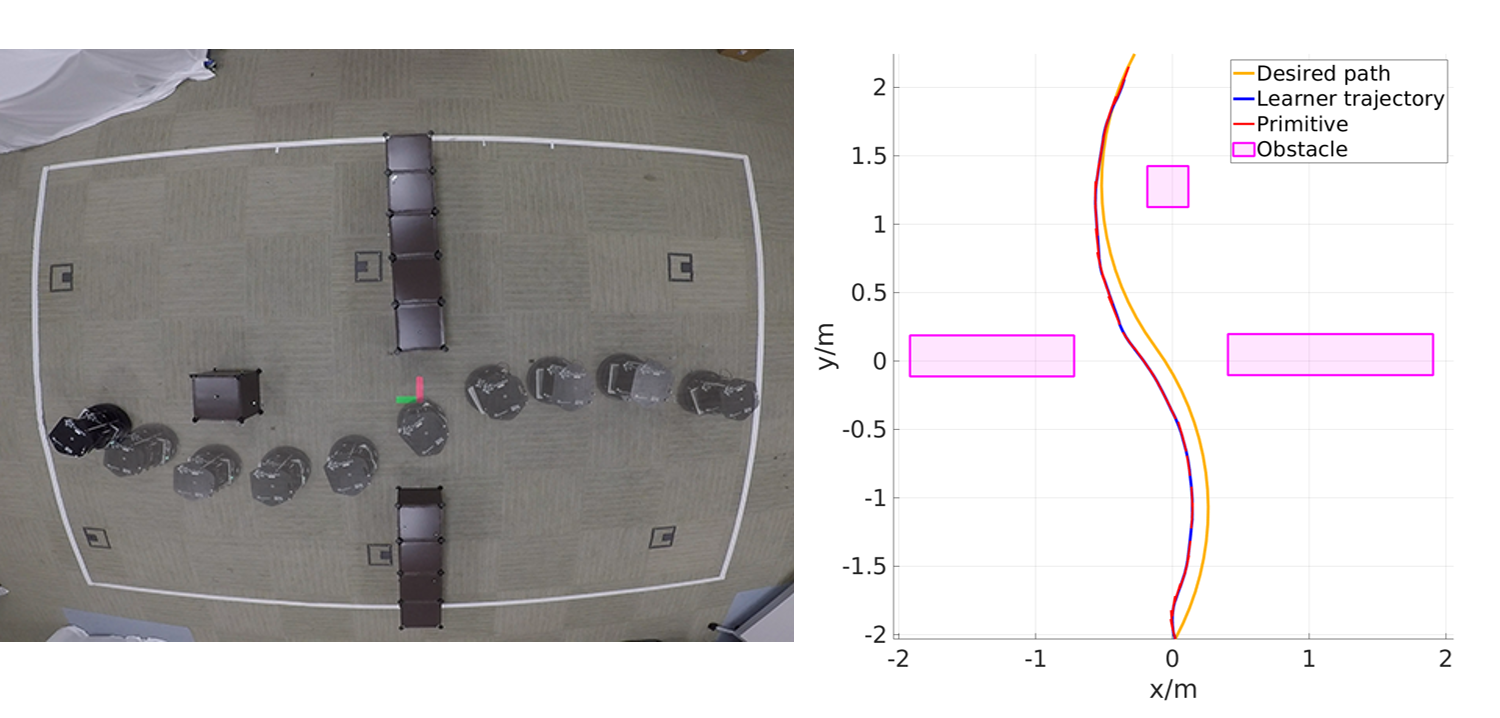}
  \vspace{-10pt}
  \caption{Turtlebot experiment with SCM.}
  \label{fig:ExpTurtlebotSCM}
\end{figure}

\section{Conclusion and future Work} \label{sec:conclusion}

In this work, we proposed a novel light-weight transfer learning framework based on conformal mapping. We use SCM to directly map the control input from the teacher to the learner without knowing the dynamical model of the learner. The framework transfers not only the control policy but also adapts the teacher's motion planning policy to make it compatible with the learner. The proposed method is validated with both simulations and actual experiments. The results show that the learner can safely adapt the control and motion planning policy to suit its own dynamics.

In our future work, we are looking into leveraging multi-dimensional conformal mapping to transfer from a higher-order system to a lower-order system, such as from an aerial vehicle to a ground vehicle. We plan also to extend our framework to deal with learners that have more capabilities than the teacher.


\section{Acknowledgements}
This work is based on research sponsored by DARPA under Contract No. FA8750-18-C-0090.

\bibliographystyle{IEEEtran}
\bibliography{References} 

\end{document}